\documentclass[
]{ceurart}

\usepackage{listings}
\usepackage{todonotes}
\usepackage[shortcuts,acronym]{glossaries}
\newacronym{ml}{ML}{Machine Learning}
\newacronym{nn}{NN}{Neural Networks}
\newacronym{fmcw}{FMCW}{Frequency-Modulated Continuous Wave}
\newacronym{rbs}{RBS}{Rule-Based System}
\newacronym{iai}{IAI}{Interpretable AI}
\newacronym{shap}{SHAP}{Shapley Additive Explanations}
\newacronym{xai}{XAI}{explainable AI}
\newacronym{er}{ER}{experience replay}
\newacronym{tl}{TL}{Transfer learning}
\newacronym{gru}{GRU}{gated recurrent unit}
\newacronym{hgr}{HGR}{hand gesture recognition}
\newacronym{mmwave}{mmWave}{millimeter-wave}
\newacronym{cnn}{CNN}{convolutional neural network}
\newacronym{lstm}{LSTM}{long short-term memory}
\newacronym{svm}{SVM}{Suport vector machines}
\newacronym{knn}{kNN}{k-nearest neighbors}
\newacronym{hmm}{HMM}{Hidden markov models}
\newacronym{pca}{PCA}{principle component analysis}
\newacronym{resnet}{ResNet}{Residual networks}
\newacronym{rmd}{RMD}{relative Mahalanobis distance}
\newacronym{gan}{GAN}{generative adversarial networks}
\newacronym{rdi}{RDI}{Range-Doppler image}
\newacronym{mira}{MIRA}{multi-class interpretable rule-based algorithm}
\newacronym{iqr}{IQR}{Interquatile range}
\newacronym{fft}{FFT}{fast Fourier transform}
\newacronym{rf}{RF}{radio-frequency}
\newacronym{if}{IF}{intermediate frequency}
\newacronym{std}{STD}{standard deviation}
\newacronym{rnn}{RNN}{recurrent neural network}
\newacronym{euai}{EU AI Act}{European Union's Artificial Intelligence Act}
\newacronym{lime}{LIME}{Local Interpretable Model-agnostic Explanations}
\newacronym{ai}{AI}{artificial intelligence}

\begin{document}

\copyrightyear{2022}
\copyrightclause{Copyright for this paper by its authors.
  Use permitted under Creative Commons License Attribution 4.0
  International (CC BY 4.0).}

\conference{xAI-2025: The 3rd World Conference on eXplainable Artificial Intelligence, July 09--11, 2025, Istanbul, Turkey}

\title{Learning Interpretable Rules from Neural Networks: Neurosymbolic AI for Radar Hand Gesture Recognition}

% \tnotetext[1]{You can use this document as the template for preparing your
%   publication. We recommend using the latest version of the ceurart style.}

\author[1,2]{Sarah Seifi}[%
orcid=0009-0003-0448-599X,
email=sarah.seifi@tum.de,
]
\cormark[1]
\address[1]{Technical University of Munich, Arcisstraße 21, 80333 Munich, Germany}
\address[2]{Infineon Technologies AG, Am Campeon 1-15, 85579 Neubiberg, Germany}
\address[3]{Johannes Kepler University Linz, Altenbergerstraße 69, 4040 Linz, Austria}
\address[4]{Software Competence Center Hagenberg GmbH (SCCH), Softwarepark 32a, 4232 Hagenberg, Austria}

\author[2,3]{Tobias Sukianto}[%
orcid=0000-0003-4136-8453,
email=tobias.sukianto@infineon.com,
]

\author[2]{Cecilia Carbonelli}[%
orcid=0000-0001-7832-7597,
email=cecilia.carbonelli@infineon.com,
]

\author[1]{Lorenzo Servadei}[%
orcid=0000-0003-4322-834X,
email=lorenzo.servadei@tum.de,
]

\author[1,4]{Robert Wille}[%
orcid=0000-0002-4993-7860,
email=robert.wille@tum.de,
]

\cortext[1]{Corresponding author.}
% \fntext[1]{These authors contributed equally.}


\begin{abstract}
Rule-based models offer interpretability but struggle with complex data, while deep neural networks excel in performance yet lack transparency. This work investigates a neuro-symbolic rule learning neural network named RL-Net that learns interpretable rule lists through neural optimization, applied for the first time to radar-based hand gesture recognition (HGR). We benchmark RL-Net against a fully transparent rule-based system (MIRA) and an explainable black-box model (XentricAI), evaluating accuracy, interpretability, and user adaptability via transfer learning. Our results show that RL-Net achieves a favorable trade-off, maintaining strong performance (93.03\% F1) while significantly reducing rule complexity. We identify optimization challenges specific to rule pruning and hierarchy bias and propose stability-enhancing modifications. Compared to MIRA and XentricAI, RL-Net emerges as a practical middle ground between transparency and performance. This study highlights the real-world feasibility of neuro-symbolic models for interpretable HGR and offers insights for extending explainable AI to edge-deployable sensing systems.
\end{abstract}

\begin{keywords}
  Interpretable Classification \sep
  Rule-Based Learning \sep
  Neuro-Symbolic AI \sep
  FMCW Radar \sep
  Hand Gesture Recognition
\end{keywords}

\maketitle

\section{Introduction}

Hand gestures offer a natural modality for human–computer interaction, with applications in automotive safety, healthcare, and augmented reality \cite{ges_auto, ges_med, ges_vr}. Radar-based \ac{hgr} is particularly attractive due to the radar’s robustness, privacy-preserving properties, and compact form factor \cite{strobel}.

Despite recent success of deep learning for radar-based HGR \cite{yan2023mmgesture}, these models are often opaque and lack interpretability, posing challenges in safety-critical and regulated environments, particularly under emerging legislation like the European Union's Artificial Intelligence Act\footnote{https://eur-lex.europa.eu/eli/reg/2024/1689/oj}.

Fully transparent, rule-based models provide interpretability but fail to generalize in complex high-dimensional settings. Neuro-symbolic \ac{ai} offers a promising middle ground by integrating symbolic logic with neural architectures \cite{self-interpretable-survey}, enabling models that learn structured, interpretable rules through gradient-based optimization.

In this work, we explore RL-Net (Rule Learning neural Network) \cite{rlnet}, a neuro-symbolic model that learns ordered rule lists for classification. We apply it to radar-based \ac{hgr} and evaluate its trade-offs between accuracy and interpretability. To our knowledge, this is the first practical application of neuro-symbolic learning to real-world \ac{fmcw} radar data, with a focus on interpretability and user adaptation.

\noindent Our main contributions are:
\begin{enumerate}
    \item \textbf{First application of RL-Net to radar-based \ac{hgr}:} We demonstrate the first real-world deployment of a neuro-symbolic model on FMCW radar gesture data.
    
    \item \textbf{Training stabilization and architectural improvements:} We enhance RL-Net with batch normalization and validation-time regularization to improve robustness and reduce rule complexity.

    \item \textbf{Personalized transfer learning:} We apply user-specific fine-tuning to RL-Net, improving accuracy while simplifying learned rule sets.
    
    \item \textbf{Comparative evaluation:} We benchmark RL-Net against MIRA (white-box) and XentricAI (black-box), assessing accuracy, interpretability, and adaptability.

    \item \textbf{Limitations and future work:} We identify structural training issues in RL-Net’s fixed rule hierarchy and outline lightweight improvements to support scalable, interpretable deployment.
\end{enumerate}

\section{Background and Related Work}
\subsection{Interpretable White-box Models}
Transparent models like decision trees, logistic regression, and rule-based systems are essential for high-stakes applications due to their interpretability \cite{frasca2024explainable}. Rule-based models, structured as "if-then" statements, align with human reasoning and are typically organized as rule sets or hierarchical rule lists \cite{skoperules}.

A rule comprises multiple feature-based conditions leading to a prediction, e.g., "IF (x1 > 0.5 AND x2 $\leq$ 3.0) THEN class = 1". This clarity supports traceability and user trust.

MIRA \cite{seifi2024mira} exemplifies a transparent rule-based classifier designed for multi-class \ac{hgr}. It incorporates foundational and user-personalized rules but suffers from limited scalability and overfitting as dimensionality increases.

\subsection{Explainable Black-box Models}
Deep \ac{nn} (convolutional \ac{nn}s \cite{ges_cnn}, recurrent \ac{nn}s \cite{strobel}, Transformers \cite{jin2023interference}) yield strong radar \ac{hgr} performance but lack interpretability. XAI techniques like SHAP \cite{shap} offer post-hoc feature attributions but are faithful to the model, not the underlying data \cite{chen2020true}. They fail to reveal intermediate reasoning steps.

Recent work such as XentricAI \cite{seifi2024xentricai} applies SHAP to recurrent NNs for gesture explanation and anomaly feedback, combining gesture detection and classification per frame. While insightful, it remains a black-box at its core.

\subsection{Neuro-Symbolic AI (Gray-box Models)}
Neuro-symbolic AI bridges rule-based interpretability and neural flexibility \cite{self-interpretable-survey}. These models generate readable rules through trainable NNs, producing interpretable outputs from opaque training, often termed "gray-box" systems \cite{li2021grey}.

DR-Net (Decision Rules Network) \cite{drnet} models rules using a two-layer \ac{nn}: an AND-based rules layer followed by an OR layer. Rules emerge from binary neuron activations with sparsity-based regularization to control complexity.

RL-Net \cite{rlnet} extends DR-Net with fixed-order rule hierarchy, producing interpretable rule lists suitable for edge deployment. More advanced systems like HyperLogic \cite{hyperlogic} improve scalability via hypernetworks but at the cost of interpretability and complexity, making them less ideal for transparent and on-the-edge applications.

\begin{figure}[h]
  \centering
  \includegraphics[width=\linewidth]{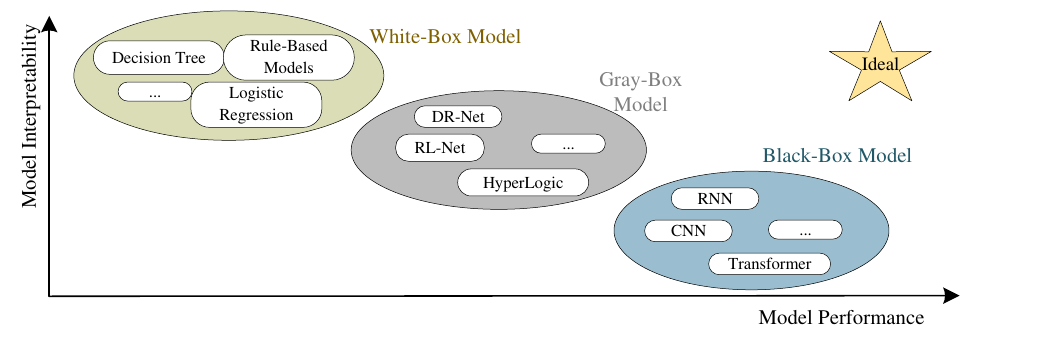}
  \caption{Comparison of model categories (white-box, gray-box, black-box) based on model complexity and predictive performance. Neuro-symbolic approaches like RL-Net occupy an intermediate "gray-box" space, balancing interpretability and performance. An ideal model has high model interpretability as well as performance.}
  \label{fig:perf-vs-interpretability}
\end{figure}

Figure~\ref{fig:perf-vs-interpretability} summarizes model trade-offs between interpretability and performance, positioning RL-Net between black-box deep models and fully symbolic approaches.

\subsection{Motivation for Our Work}
To date, neuro-symbolic models have not been applied to real-world FMCW radar gesture data. Their adaptability via \ac{tl} also remains underexplored. This work fills that gap by evaluating RL-Net in comparison to MIRA and XentricAI, analyzing the performance–interpretability trade-off and exploring its potential for user-specific gesture adaptation.

\section{Methodology}

\subsection{RL-Net Overview}

\begin{figure}[h]
\centering
\includegraphics[width=\linewidth]{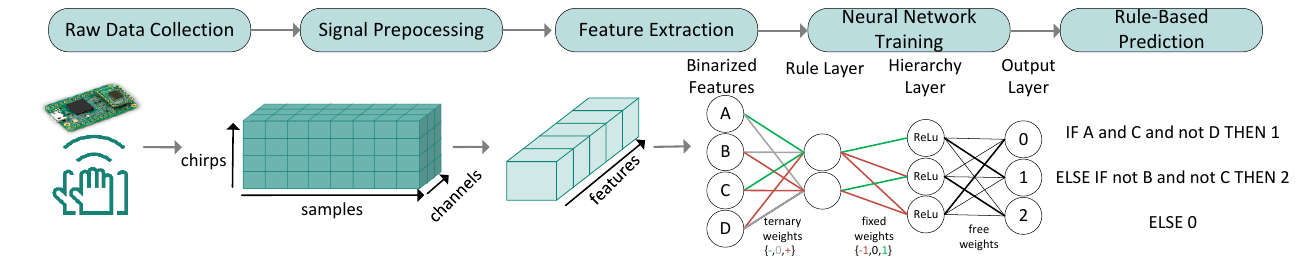}
\caption{Overview of the proposed interpretable gesture recognition pipeline.}
\label{fig:pipeline_overview}
\end{figure}

We apply RL-Net \cite{rlnet} to \ac{fmcw} radar gesture data. The pipeline involves signal preprocessing, feature extraction, model training, and interpretable rule generation, as illustrated in Figure \ref{fig:pipeline_overview}. 

\textbf{Input \& Rule Layers.} The input layer receives binarized features \(x_i \in \{0,1\}\). Each rule neuron functions as a logical AND, with ternary weights \(\mathbf{W}_T\) encoding positive, negative, or excluded feature use. To enable differentiable training, these weights are reparameterized as \(\mathbf{W}_T = \mathbf{W}_S \circ \mathbf{W}_H\), where \(\mathbf{W}_H \in [0,1]\) are approximated binary masks following \cite{louizos2018learning}.

Each neuron computes pre-activation $y_r$
\[
y_r = \sum_i w_i x_i - \sum_{w_i > 0} w_i + 1.
\]

To improve training stability, batch normalization is applied to $y_r$ before binarization via $b(y_r)=1[y_r=1]$.

\textbf{Hierarchy \& Output Layers.} A fixed hierarchy enforces ordered rule evaluation: a rule \(R_k\) activates only if all previous rules are inactive. The final neuron acts as a default rule. Only one rule fires per sample, and a softmax in the output layer assigns its associated class label.

\textbf{Loss Function.} The regularization loss is defined as $ \mathcal{L}_\text{sparse} = \sum_{i,j} \sigma\left( \frac{\text{loc}_{ij} - \gamma}{\zeta - \gamma} \right) $, where \( \text{loc}_{ij} \) are the logits controlling the binary mask sampling, and \( \gamma, \zeta \) are stretch parameters controlling the thresholding behavior. This penalty encourages many weights to converge to zero and thus shortens the learned rules. The training optimizes the loss based on the cross-entropy loss, the sparsity and L2 regularization terms:
\[
\mathcal{L} = \mathcal{L}_{\text{CE}} + \lambda_1 \mathcal{L}_{\text{sparse}} + \lambda_2 \|\mathbf{W}_{\text{out}}\|_2^2
\]

\subsection{Comparison to Baselines}

We compare RL-Net to MIRA \cite{seifi2024mira}, a white-box rule-based system using foundational and user-specific rules, and XentricAI \cite{seifi2024xentricai}, a black-box GRU model explained post-hoc via SHAP.

A summary of capabilities is shown in Table~\ref{tab:method_comparison}.

\begin{table}[h]
\centering
\small
\caption{Comparison of explainable and interpretable \ac{hgr} approaches.}
\label{tab:method_comparison}
\begin{tabular}{lccc}
\toprule
\textbf{Method} & \textbf{Explainability Type} & \textbf{Adaptability} & \textbf{Scalability} \\\midrule
MIRA \cite{seifi2024mira}      &  White-box (fully interpretable)  & High (personalized rules)     & Low \\
XentricAI \cite{seifi2024xentricai}  &  Black-box (post-hoc explanations)     & High (\ac{tl})       & High \\
RL-Net \cite{rlnet}     &  Gray-box (semi-interpretable)     & Medium–high (\ac{tl}) & Medium–high \\
\bottomrule
\end{tabular}
\end{table}

\section{Dataset and Preprocessing}
\subsection{FMCW-Radar Gesture Dataset}
The gesture dataset used in this study was collected using Infineon Technologies' XENSIV™ BGT60LTR13C $60\,$GHz FMCW radar operating in indoor environments. Twelve users performed five predefined hand gestures (\textit{SwipeLeft}, \textit{SwipeRight}, \textit{SwipeUp}, \textit{SwipeDown}, and \textit{Push}) across six different room types. Each participant completed $1{,}000$ samples, totaling $31{,}000$ gesture recordings.

Recordings were captured over $100$ frames per gesture using three receive antennas, with dimensions $[100 \times 3 \times 32 \times 64]$ representing frames, antennas, chirps, and fast-time samples, respectively. The dataset is publicly available via IEEE Dataport \cite{radar_data}.

\subsection{Frame-Level Gesture Labeling}
Following the method in \cite{strobel}, we assign gesture labels based on the frame with minimum radial distance (gesture anchor point). A fixed 10-frame window centered around this point defines the gesture duration, while all other frames are labeled as background. This results in a labeled dataset with dimensions $[M \times F \times D]$, where $M$ is the number of samples, $F$ the number of frames per sample, and $D = 5$ the number of features.

\subsection{Signal Processing and Feature Extraction}
We applied a lightweight, real-time-capable preprocessing pipeline based on \cite{strobel} to extract gesture-relevant features from the raw radar data. The main steps include: (i) \textbf{Range FFT}, applied after DC removal to generate range profiles; (ii) \textbf{Moving Target Indication (MTI)}, which removes static target reflections; (iii) \textbf{Target Localization}, using peak detection in the range profile to find the closest moving object (i.e., the hand); (iv) \textbf{Doppler FFT}, applied to the identified hand bin to extract radial velocity; and (v) \textbf{Angle Estimation}, computing azimuth and elevation from antenna phase differences.

From this processing, we extract five features per frame: radial distance (range), radial velocity (Doppler), azimuth angle, elevation angle, and signal magnitude. These were averaged over the ten gesture frames to construct the input to the models.

\section{Experimental Setup}

To evaluate both the generalization capabilities and adaptability of RL-Net, we conducted two main experiments: (i) optimized training of a user-agnostic pretrained model, and (ii) user-specific \ac{tl}. Gesture data from twelve users (12,000 samples) was used in total. Data from six users (6,000 samples) was used to train and validate the base model, while data from the remaining six users was reserved for \ac{tl} and evaluation.

\textbf{Dataset Split.} Six users (6,000 gestures) were used to pretrain all models; six different users (6,000 gestures) were held out for \ac{tl} and testing.

\textbf{Training Protocol.} An extensive hyperparameter search was performed using Grid Search, selecting values based on prior work and empirical stability. We used the Adam optimizer with a learning rate of 0.01 and batch size of 40. Training was run for 200 epochs with early stopping on validation loss. A batch normalization layer was added after the rule layer for improved stability. Sparsity was controlled via a regularization weight (\(\lambda_{\text{1,train}} = 0.025\)), while a higher weight (\(\lambda_{\text{1,val}} = 0.3\)) was applied during validation to promote simpler models.  L2 regularization was disabled. Hard concrete parameters were \(\gamma = -0.1\), \(\zeta = 1.1\), and \(\beta = \frac{2}{3}\).

\textbf{Custom Early Stopping Strategy.} Standard early stopping based on validation cross-entropy often favored overly complex models with minimal gains in accuracy. On the other hand, strong regularization led to premature convergence and vanishing gradients. To balance interpretability and performance, we introduced a custom early stopping criterion using an increased validation-time regularization weight (\(\lambda_{1,\text{val}} = 0.3\)). This steered model selection toward sparser rule sets without compromising predictive accuracy.

\textbf{Transfer Learning.} For user adaptation, pretrained weights were partially frozen: rule masks and output weights were updated while base rule weights were fixed. Each user had 1,000 samples (20\% test, 64\% train, 16\% val).

\textbf{Evaluation Metrics.} We report accuracy and F1 score, along with rule complexity measured by the number of active rules and total rule conditions. These metrics are tracked before and after fine-tuning.

\textbf{Baselines.} RL-Net was compared to MIRA \cite{seifi2024mira}, and XentricAI \cite{seifi2024xentricai}. Note that XentricAI performs both gesture detection and classification by explicitly modeling a background class, so comparisons should consider this broader output scope. For both baselines, we adopt the hyperparameters and training protocols reported in their original publications.

\section{Experimental Results and Discussion}
\subsection{General Training and Model Robustness}

While we initially experimented with DR-Net by training independent rule sets for each gesture class, the resulting rules were not mutually exclusive, making them unsuitable for robust multi-class classification. As a result, we focused exclusively on RL-Net, which provides a hierarchical rule list enabling consistent multi-class predictions.

We evaluate the model’s robustness and interpretability across three architectural variants: the original RL-Net, RL-Net with batch normalization, and RL-Net with both batch normalization and a modified validation loss incorporating regularization weight \(\lambda_{1,\text{val}}\). Results, averaged over ten runs per configuration with consistent seeding, are summarized in Table \ref{tab:arch_comparison}.

\begin{table}[h]
\centering
\small
\caption{Performance and complexity across RL-Net architectural variants (mean ± 95\% confidence interval).}
\label{tab:arch_comparison}
\begin{tabular}{lcc}
\toprule
\textbf{Architecture} & \textbf{F1 Score [\%]} & \textbf{ Conditions} \\
\midrule
Original              & 84.14 ± 2.38           & 240.45 ± 41.07              \\
+ BatchNorm           & \textbf{90.44 ± 0.95}           & 74.70 ± 32.83               \\
+ BatchNorm + \(\lambda_{1,\text{val}}\) & 90.08 ± 1.16           & \textbf{55.6 ± 17.49}      \\
\bottomrule
\end{tabular}
\end{table}

Introducing batch normalization significantly improved both performance stability and model simplicity. Further incorporating validation-time regularization reduced model complexity even more, with only a slight trade-off in F1 score. The reduction in standard deviation across runs also indicates improved training consistency.

\paragraph{Optimization Bottleneck from Rule Ordering and Vanishing Gradients.} 

\begin{figure}[h]
  \centering
  \includegraphics[width=\linewidth]{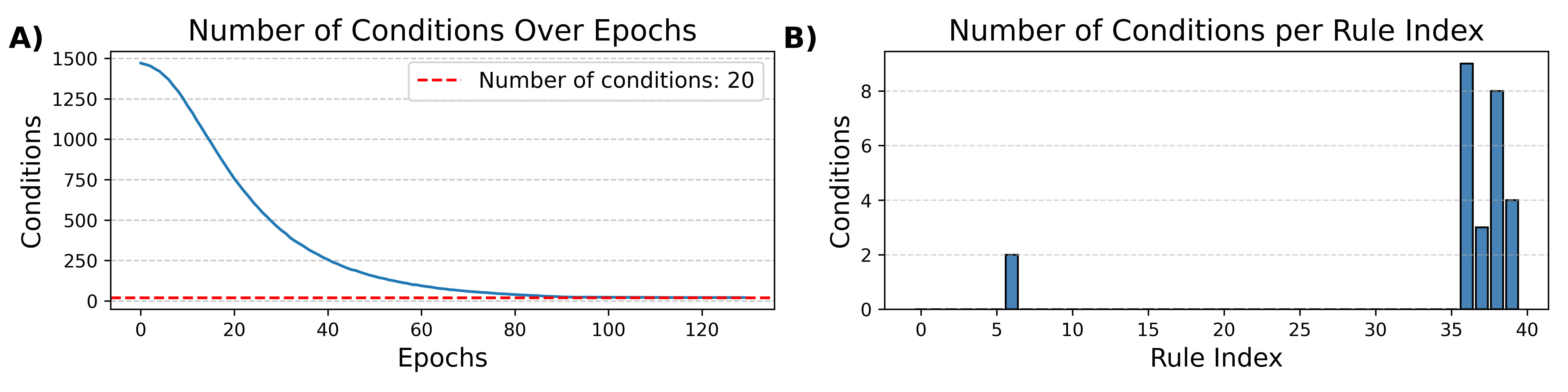}
  \caption{Training dynamics of RL-Net. \textbf{(A)} Total number of active rule conditions over training epochs, showing progressive sparsity and improved interpretability. \textbf{(B)} Histogram of surviving rule indices and their number of conditions in a representative run. Due to the fixed rule evaluation hierarchy, early rules are pruned, while later rules dominate, highlighting a structural imbalance.}
  \label{fig:rule_evolution}
\end{figure}

While training improves overall sparsity (Figure~\ref{fig:rule_evolution}, Panel A), we observed a persistent structural issue: RL-Net's fixed hierarchy layer causes early rules to being suppressed and pruned. As shown in Panel B, this results in only high-index rules remaining active, often with long, complex conditions, compromising rule diversity and interpretability. This bottleneck stems from the fixed top-down rule evaluation, which prioritizes later rules during optimization. Addressing this may require adaptive or learnable rule ordering, ideally without increasing model complexity to maintain suitability for edge deployment.

\paragraph{Comparison with Baselines.} As a reference, the white-box model MIRA achieves an F1 score of 79.7\%, while the GRU-based classification backbone of XentricAI reaches 95\%. RL-Net demonstrates a compelling trade-off: It achives strong performance (\(\sim\)90\%) while maintaining interpretable, compact rule lists. This positions RL-Net effectively between the extremes of full transparency and pure black-box modeling.

\subsection{Transfer Learning}

For these experiments, we initialized the model using a randomly selected baseline model consisting of seven rules and 36 total conditions.

Table~\ref{tab:transfer_learning_results} summarizes the classification performance and rule complexity before and after \ac{tl}. All results are reported for the best-performing epoch on the validation set, using accuracy and F1-score as primary metrics, and the number of rules and total rule conditions as proxies for interpretability.

\begin{table}[h]
\centering
\small
\renewcommand{\arraystretch}{0.7} % Slightly tighter row spacing
\caption{TL results per user: Accuracy (Acc), F1-score (F1), and rule complexity.}
\label{tab:transfer_learning_results}
\begin{tabular}{ccccccc}
\toprule
\textbf{User} & \textbf{Initial Acc [\%]} & \textbf{Initial F1 [\%]} & \textbf{TL Acc [\%]} & \textbf{TL F1 [\%]} & \textbf{\#Rules} & \textbf{\#Conditions} \\
\midrule
1 & 84.5 & 84.52 & 95.5 & 95.46 & 5 & 29 \\
2 & 91.0 & 91.11 & 93.0 & 92.97 & 5 & 11 \\
3 & 89.5 & 89.50 & 90.5 & 90.21 & 5 & 30 \\
4 & 95.3 & 95.15 & 95.8 & 95.81 & 5 & 36 \\
5 & 94.0 & 94.01 & 95.0 & 95.02 & 5 & 18 \\
6 & 85.0 & 84.18 & 91.0 & 90.91 & 5 & 36 \\
\midrule
\textbf{Avg.} & \textbf{89.9} & \textbf{88.75} & \textbf{93.05} & \textbf{93.06} & \textbf{5} & \textbf{26.67} \\
\bottomrule
\end{tabular}
\end{table}

Across all users, \ac{tl} consistently improved model performance, with F1-score increases ranging from modest (\(+0.71\%\) for user$_3$) to substantial (\(+9.08\%\) for user$_1$). Importantly, these gains were accompanied by a consistent reduction in rule complexity. All adapted models converged to five rules, and in several cases, the total number of rule conditions decreased significantly, highlighting the dual benefit of enhanced personalization and improved interpretability.

\paragraph{Comparison with Baselines.} RL-Net achieved an average user-specific performance of 93.05\%, surpassing the fine-tuned XentricAI model, which reached 90.2\%. However, it is important to note that XentricAI includes a \textit{Background} class and performs both gesture detection and classification at the frame level, an extended functionality that goes beyond RL-Net’s current scope.
MIRA achieved the highest average accuracy of 94.9\% after user-specific calibration. While deterministic, MIRA relies on handcrafted rule tuning and lacks the learning flexibility of neural approaches.

Overall, RL-Net strikes a favorable balance between interpretability and performance, offering greater modeling flexibility than MIRA and better classification accuracy than XentricAI. That said, one limitation may lie in the fixed thresholds used for binarizing continuous features, which could constrain the model’s adaptability to subtle user-specific variations. Future research should explore adaptive thresholding or learnable binarization strategies to further improve generalization and personalization.

\subsection{Limitations and Future Work}

RL-Net inherits optimization challenges from DR-Net, notably vanishing gradients caused by multiplicative logical activations. This leads to early-rule pruning and dominance of later, often longer rules, limiting generalization. Fixed rule ordering further amplifies this by prioritizing high-indexed neurons and suppressing earlier ones.

While batch normalization improved stability, training remains inconsistent across runs. Attempts like L2 regularization worsened gradient flow and performance. Though more advanced techniques (e.g., HyperLogic \cite{hyperlogic}) exist, they add model complexity, compromising interpretability and edge-suitability.

Future work should explore gradient-stable activations, adaptive thresholding for input binarization, and learnable rule hierarchies. Physics-informed constraints tailored to gesture kinematics may further improve robustness without added complexity. Additionally, a user study could be incorporated to better evaluate and refine the interpretability of the model in practical, real-world scenarios. These insights extend beyond \ac{hgr}, pointing to broader improvements in neuro-symbolic learning.

\section{Conclusion}

This work presents the first real-world application of RL-Net, a neuro-symbolic model, to radar-based \ac{hgr}. RL-Net achieves a strong balance between interpretability and performance, outperforming fully transparent models and approaching the accuracy of black-box methods.

Through optimized training and \ac{tl}, we demonstrate RL-Net’s potential for user-adaptive, explainable gesture sensing. However, challenges in optimization and rule hierarchy remain. Addressing these without increasing model complexity is key to advancing interpretable AI for edge deployment and beyond.

%% Define the bibliography file to be used
\bibliography{sample-ceur}

\appendix

\end{document}